# Minimally Actuated Serial Robot

Moshe P. Mann, Lior Damti, David Zarrouk

*Abstract*— In this paper, we propose a novel type of serial robot with minimal actuation. The robot is a serial rigid structure consisting of multiple links connected by passive joints and of movable actuators. The novelty of this robot is that the actuators travel over the links to a given joint and adjust the relative angle between the two adjacent links. The joints passively preserve their angles until one of the actuators moves them again. This actuation can be applied to any serial robot with two or more links. This unique configuration enables the robot to undergo the same wide range of motions typically associated with hyper-redundant robots but with much fewer actuators. The robot is modular and its size and geometry can be easily changed. We describe the robot's mechanical design and kinematics in detail and demonstrate its capabilities for obstacle avoidance with some simulated examples. In addition, we show how an experimental robot fitted with a single mobile actuator can maneuver through a confined space to reach its target.

## I. INTRODUCTION

Hyper redundant robots are robots with serially connected links that possess a large kinematic redundancy. Alternatively known as snake robots, they are the subject of extensive research over the past several decades [1] [2] [3].

with many different configurations, mechanisms, control strategies, and motion planning algorithms being proposed over the years. The principle motivation for developing hyper redundant robots is their ability to navigate around obstacles and in highly confined spaces.

Algorithms for planning the motion of hyper redundant robot present a formidable challenge [4] [5]. Early motion planners for hyper-redundant robot motion planning were developed by Gregory Chirkjian in [6] [7] [8] [9]. In those works, the curvature of the robotic snake was approximated as a continuous modal function with the obstacles expressed as boundary constraints on the robot's shape. Many recent works have addressed obstacle avoidance schemes for hyper redundant robots. State-of-the-art approaches including genetic algorithms [10] [11], variational methods [12], and probabilistic roadmaps [13] are used to plan the motions of the robots. There is a continuous progress in reducing the planning time and improving their capability in real life scenarios such as robotic surgery, agriculture and search and rescue.

In parallel, flexible robots have been developed as an alternative. Also known as soft robots or continuum robots, they consist of a flexible continuous structure that possess, at least in theory, an infinite number of degrees of freedom. The advantage of flexible robots over hyper-redundant robots is their lightweight and speed. However, there is still ongoing research to improve their accuracy, control and position and sensing capabilities (see [14] and [15]).

In this work, we propose the Minimally Actuated Serial Robot (MASR) which combines some characteristics and advantages from both hyper redundant robots and compliant robots. The MASR is a serial robot consisting of multiple links connected by passive joints and of a small number of movable actuators. The actuators translate over the links to any given joint and adjust it to the desired angular displacement. The joint passively preserves its angle until it is actuated again. The number of degrees of reconfigurability (DOR) is equal to the number of joints. This enables the MASR to achieve similar mobility (albeit slower) to regular hyper redundant robots. The advantages of MASR are its simplicity, smaller weight, higher energy density (power/mass), low cost and modularity, as the number of links and actuators can be easily and quickly changed.

We describe the mechanism of the MASR in Section II. In Section III, the kinematics of the robot are outlined. Section IV provides some examples of motion planning around obstacles that the MASR achieves. In Section V, we demonstrate how the MASR can duplicate the motion of a fully actuated hyper-redundant robot to any desired degree of accuracy. Several examples of this are given in Section VI using multiple links and single mobile actuator. Conclusions and directions for further research are given in Section VII.

## II. MECHANISM DESCRIPTION AND KINEMATICS

Our novel robot system is composed of $N$ links connected through passive joints, $M$ mobile actuators that travel over the links, and an end effector as shown in Figure 1. The passivity of the joints is defined by there being no motors in between them, while the angle between adjacent links is preserved. The number of links and mobile actuators can be easily varied depending on the proposed task. When a mobile actuator travels over the links, it can rotate the desired joint thereby changing the relative angle between the links by a desired angle. The base is where the robot is connected to a constant support or a mobile platform.

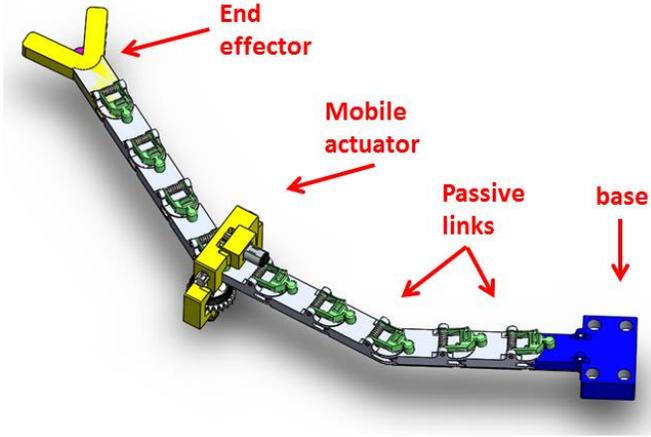

Figure 1. A 2D prototype of the Minimally Actuated Robotic Snake. The robot in this figure has 10 links, one mobile actuator, and an end effector. The mobile actuator can freely travel over the links and rotates them upon command.

For simplicity, we assume that each link is of uniform length $L$. The angle between the $i$-1th and $i$th serial link is denoted by $\theta_i$.

The orientation of each link in world coordinates is $\alpha_i$

$$\alpha_i = \sum_{l=1}^{i} \theta_l \quad (1)$$

and its position is given by:

$$(x_i, y_i) = \left( L\sum_{k=1}^{i}\cos\left(\sum_{l=1}^{k}\theta_l\right), L\sum_{k=1}^{i}\sin\left(\sum_{l=1}^{k}\theta_l\right) \right) \quad (2)$$

The coordinate of the $j$th actuator is given by the pair $(n_j, \theta_j) \in \mathbb{N} \times \mathbb{I}$, where $n_j$ is the link at which actuator $j$ is currently located and $\theta_j$ is the angle of the actuator *and* the joint that the actuator is currently actuating, being that the latter two must be equal. The actuator angle has the same range as $\theta$. We denote the set of actuated joints as $J_A$ and the set of unactuated joints as $J_U$, given formally by

$$\begin{aligned} J_A &= \{n_1, \ldots n_M\} \\ J_U &= \{1, \ldots N\} \setminus J_A \end{aligned} \quad (3)$$

The configuration space of the robot, assuming there are joint limits, is an $N$ dimensional cube $\mathbb{I}^N$, where $\mathbb{I}$ is open one dimensional ball. However, the reduced actuation of the serial robot results in a very significant kinematic constraint. For any given set of actuator locations $n_1, n_2, \ldots n_M$, the motion of the robot is confined to an $M$ dimensional manifold embedded in $\mathbb{I}^N$. This manifold is an $M$ dimensional plane in the coordinate space spanned by the unit vectors $\mathbf{e}_{n_1}, \mathbf{e}_{n_2}, \ldots \mathbf{e}_{n_M}$ passing through a point $\mathbf{p}_u = [p_1, \ldots p_N]^T \in \mathbb{I}^N$ given by:

$$p_i = \begin{cases} 0 & i \in J_A \\ \theta_i & i \in J_U \end{cases} \quad (4)$$

The constraint on the set of joint angles $\boldsymbol{\theta} = [\theta_1, \theta_2, \ldots, \theta_N]$ in c-space is thus expressed as:

$$\boldsymbol{\theta} \in \mathbf{p}_u + \sum_{a \in J_A} k_a \mathbf{e}_{n_a}, \quad k_a \in (-2\pi, 2\pi) \quad (5)$$

Translating the actuators of the robot thus corresponds to moving the manifold to a different plane in coordinate space. This has significant ramifications for motion planning, as will become apparent in Section III. The trajectory of the robot through configuration space is given by the parametrized curve $f : [0,1] \to \mathbb{I}^N$ that is not $C^1$ continuous.

The total time $t_{total}$ required for the robot to reach a goal is thus comprised of the times required to rotate each joint plus the times required to traverse the actuator from one link to another plus a certain interruption delay between the translation and rotation. The latter two are a consumption of time unique to the MASR robot, and it is the price we pay for using less actuators than joints – there must be a "timeshare" of the actuator between the links.

If we assume constant translational speed $V$ of the mobile actuator and constant rotational speed $\omega$, and that the delay is $T_{delay}$, then the time required to perform a task is:

$$T_{TOTAL} = \frac{L\sum_{i}^{N_{STEP}}|\Delta n_i|}{V} + \frac{\sum_{i}^{N_{STEP}}|\Delta \theta_i|}{\omega} + N_{STEP} * T_{DELAY} \quad (6)$$

where $N_{STEP}$ is the total number of steps, and $\Delta n_i$ and $\Delta \theta_i$ are the respective number of links and rotation traversed during step $i$. The total energy $E$ consumed by the MASR, assuming a linear model of energy consumption dependence on coordinate displacement, would be proportional to the number of actuator translations and joint displacements:

$$E_{TOTAL} = k_n L \sum_{i}^{N_{STEP}}|\Delta n_i| + k_\theta \sum_{i}^{N_{STEP}}|\Delta \theta_i| \quad (7)$$

where $k_n$ and $k_\theta$ are coefficients that can be determined empirically. These traversals of the actuator are the more time consuming action, and it would therefore be desirable to minimize the number of traversals. This would constitute a suitable optimization goal of any motion planning algorithm for the MASR robot, and is the subject of ongoing research.

### III. FULLY ACTUATED MOTION DUPLICATION

The minimal actuation of the serial robot means that its motion is more limited than that of a fully actuated serial robot. The motions executed by a fully actuated robot cannot be completely mimicked by the MASR. However, one may desire to approximate the motions of a fully actuated robot with an MASR to within a certain degree of accuracy. In approximating the motion of the robot, there are two possible general objectives: to approximate the motion of the end effector in the work space, or to approximate the motion of the robot in coordinate space, or c-space for short – i.e. the joint angles.

The former objective seems to be the more convenient and useful goal, as the positioning of the end-effector is what defines the accuracy of the task for many robotic applications. However, the constraints on motion, expressed by Equation (5), are on the joint angles. Therefore, it is more straightforward to express error bounds on the joint angles of the robots than on the end-effector. This error bound is denoted by $\delta$, and is used as a measure of the closeness of approximation of the MASR robot to a fully actuated robot. To this end, we formulate the following definition:

**Definition 1:** A curve $f(t)$ is a $\delta_p$ *approximation* of a curve $g(u)$ if for all $t \in [0,1]$, there exists $u \in [0,1]$ and for all $u \in [0,1]$, there exists $t \in [0,1]$ such that $\|f(t) - g(u)\|_p \leq \delta$.

In other words, if all points along the trajectory of the MASR in *c*-space are close to at least some point along the trajectory of the fully actuated robot and vice versa – i.e. within a "sphere" of radius $\delta$, then the motion of the robot is sufficiently approximated. Although any norm can be used to define the sphere, we select the $\infty$-norm. This means that if we denote the *i*th dimension of a point $g(u)$ as $g^i(u)$, then:

$$\max_{i=1,\ldots,N} |f^i(t) - g^i(u)| \leq \delta \qquad (8)$$

Denoting $f(t) = \theta(t)$ as representing the configuration of the MASR and $g(u) = \theta_0(t) = [\theta_{10}, \theta_{20}, \ldots, \theta_{N0}]$ that of the fully actuated robot, then Equation (8) is equivalent to:

$$|\theta_i - \theta_{i0}| \leq \delta \quad \forall i \in \{1, N\} \qquad (9)$$

The reason for this selection is because the constraint of Equation confine the joint angles to an *M*-dimensional plane spanned by *M* vectors parallel to *M* out of *N* axis. This plane is by definition coincident with the surface of an *N*-dimensional cube. Equation (8), which uses the $\infty$-norm, defines a cube in *N*-dimensional space, and therefore the $\infty$-norm is the most natural one to use.

One might ask if it is kinematically possible for the minimally actuated robot to approximate the motions of a fully actuated robot in any arbitrary configuration space and to any degree of accuracy. The answer is yes:

**Lemma 1:** For all $g(u)$, there exists a $\delta_p$ approximation for all $\delta > 0$.

Such an approximate curve can be constructed for the $p=\infty$ norm using the procedure APPROXIMATION-CURVE. A flowchart of APPROXIMATION-CURVE is shown in Figure 2.

*Procedure* APPROXIMATION-CURVE

*Inputs*:
- a $C^0$ curve $g$ in $N$ dimensional space $g : [0,1] \to \mathbb{R}^N$
- curve parameter $u \in [0,1]$
- number of actuators $M$
- error norm $\delta > 0$

*Output*: an array of *N*-dimensional points *[x(1) x(2)… x(end)]* representing the path of the MASR which traverses in a straight line in coordinate space from *x(j)* to *x(j+1)*

1. Start at $u_0 = 0$ and $x(1) = g(0)$.
2. Using a nonlinear equation solver, find the lowest $u > u_0$ for which $\|g(u) - g(u_0)\|_\infty = \frac{1}{2}\delta$. Label this $u_e$.
3. Construct an *N* dimensional hypercuboid spanned by corners $g(u_0)$ and $g(u_e)$. The hypercuboid is described by the set of all points $x$ such that
$$max(g^i(u_e), g^i(u_0)) \geq x^i \geq min(g^i(u_e), g^i(u_0))$$
4. Find the shortest path between $g(u_0)$ and $g(u_e)$ along an *M* dimensional surface of the hypercuboid. Such techniques for finding the shortest path are outlined in [17]. This path will consist of straight lines connecting *N-M* vertices between $g(u_0)$ and $g(u_e)$.
5. Append these vertices *[y(1) y(2)… y(N-M) $g(u_e)$]* to the end of *[x]*.
6. Set $u_0 = u_e$.

**Proof**:
From Step 2, the metric distance between $g(u_0)$ and all $g(u)$ from $u_0$ and $u_e$ is less than or equal to $\delta/2$. Step 4 constructs the portion of $f(t)$ that lies on a surface of a hypercuboid between corners $g(u_0)$ and $g(u_e)$. Label the ends of the domain of this portion $t_0$ and $t_e$, respectively. Because all points in the hypercuboid have a metric distance of less than $\|g(u_e) - g(u_0)\|$ from any of its corner's as Step 3 describes, we have:

$$\|f(t) - g(u_0)\| \leq \|g(u_e) - g(u_0)\| = \frac{1}{2}\delta, \quad t \in [t_0, t_e] \quad (10)$$

Using the triangle inequality for normed metric spaces and applying the result of Equation (10) yields:

$$\|f(t)-g(u)\| \leq \|f(t)-g(u_0)\| + \|g(u_0)-g(u_e)\|$$
$$\leq \frac{1}{2}\delta + \frac{1}{2}\delta = \delta, \quad t \in [t_0, t_e], u \in [u_0, u_e]$$
(11)

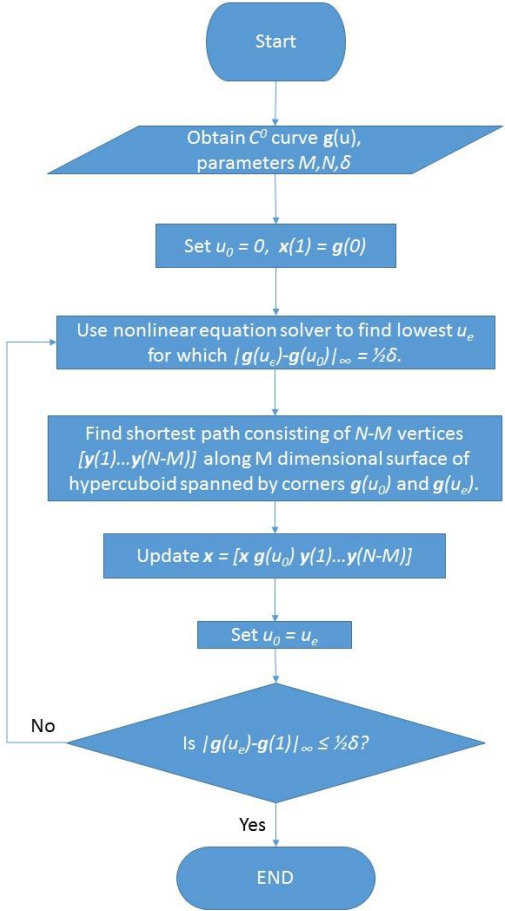

Figure 2. Flowchart of APPROXIMATION-CURVE. This algorithm constructs the trajectory of the MASR robot to track a fully actuated robot while adhering to the motion constraints of M actuators.

Step 5-7 construct $f(t)$ continuously for the entire domain of $t$. The inequality of Equation (11) thus holds for all of $f(t)$ and therefore satisfies Definition 1. □

The resulting path is clearly $C^1$ discontinuous. It goes without saying that the closer the approximation is, the more times the actuator will have to translate between links, thereby lengthening the time consumed.

## IV. ERROR ANALYSIS

While the aforementioned approximation procedure preserves the motion of the MASR within the joint error in c-space, the error of the robot endpoint in the robot workspace is in general the ultimate concern. This leads to the question: How can we determine the bound on the error of the end effector given $\delta$? In other words, if the absolute deviation of each angle of the MASR from the fully actuated robot is less than or equal to $\delta$, $x_e$ is the endpoint of the MASR, and $x_{e0}$ is the endpoint of the corresponding fully actuated robot, then what is

$$\max_{t \in [0,1]} \|x_e(t) - x_{e0}(t)\|_2 \leq f(\delta) \quad (12)$$

where $f(\delta)$ is an explicit formula relating the angular deviation $\delta$ to the 2-norm of the endpoint deviation? To calculate this dependence, we rewrite the position of the endpoint by combing Equations (1) and (2) as:

$$(x_e, y_e) = \left( L \sum_{i=1}^{N} \cos(\alpha_i), L \sum_{i=1}^{N} \sin(\alpha_i) \right)$$
$$(x_{e0}, y_{e0}) = \left( L \sum_{i=1}^{N} \cos(\alpha_{i0}), L \sum_{i=1}^{N} \sin(\alpha_{i0}) \right)$$
(13)

where $\alpha_i$ is the orientation of the $i$th joint of the MASR and $\alpha_i$ of the fully actuated robot. Using Equation (13) to express the error norm between the two endpoints yields:

$$\|x_e - x_{e0}\|_2 = L \left\| \sum_{k=1}^{i} \cos(\alpha_i) - \cos(\alpha_{i0}), \sum_{k=1}^{i} \sin(\alpha_i) - \sin(\alpha_{i0}) \right\|_2$$
$$= L \sqrt{\left( \sum_{k=1}^{i} \cos(\alpha_i) - \cos(\alpha_{i0}) \right)^2 + \left( \sum_{k=1}^{i} \sin(\alpha_i) - \sin(\alpha_{i0}) \right)^2}$$
(14)

By making use of the trigonometric identities

$$\sin A - \sin B = 2 \cos \frac{A+B}{2} \sin \frac{A-B}{2}$$
$$\cos A - \cos B = -2 \sin \frac{A+B}{2} \sin \frac{A-B}{2}$$
(15)

the terms inside the root symbol of Equation (14) become

$$\left( \sum_{i=1}^{N} \cos(\alpha_i) - \cos(\alpha_{i0}) \right)^2 + \left( \sum_{i=1}^{N} \sin(\alpha_i) - \sin(\alpha_{i0}) \right)^2$$
$$= 4 \left( \sum_{i=1}^{N} \sin \frac{1}{2}(\alpha_i + \alpha_{i0}) \sin \frac{1}{2}(\alpha_i - \alpha_{i0}) \right)^2$$
$$+ 4 \left( \sum_{i=1}^{N} \cos \frac{1}{2}(\alpha_i + \alpha_{i0}) \sin \frac{1}{2}(\alpha_i - \alpha_{i0}) \right)^2$$
(16)

Rearranging Equation (16), inserting the result into Equation (14), and squaring yields:

$$\|x_e - x_{e0}\|^2 = 4L^2 \sum_{i=1}^{N} sin^2 \frac{1}{2}(\alpha_i - \alpha_{i0}) * ... \quad (17)$$

$$\left( sin^2 \frac{1}{2}(\alpha_i + \alpha_{i0}) + cos^2 \frac{1}{2}(\alpha_i + \alpha_{i0}) \right)$$

$$+ 4L^2 \sum_{i=1}^{N} \sum_{j=1, j \neq i}^{N} \begin{pmatrix} sin \frac{1}{2}(\alpha_i - \alpha_{i0}) sin \frac{1}{2}(\alpha_j - \alpha_{j0}) * ... \\ sin \frac{1}{2}(\alpha_i + \alpha_{i0}) sin \frac{1}{2}(\alpha_j + \alpha_{j0}) + ... \\ cos \frac{1}{2}(\alpha_i + \alpha_{i0}) cos \frac{1}{2}(\alpha_j + \alpha_{j0}) \end{pmatrix}$$

Using the trigonometric identities

$$cos^2 \theta + sin^2 \theta = 1 \quad (18)$$
$$cos(A - B) = cos A cos B + sin A sin B$$

Equation (17) becomes:

$$\|x_e - x_{e0}\|^2 = 4L^2 \sum_{i=1}^{N} sin^2 \frac{1}{2}(\alpha_i - \alpha_{i0})...$$
$$+ 4L^2 \sum_{i=1}^{N} \sum_{j=1, j \neq i}^{N} \frac{sin \frac{1}{2}(\alpha_i - \alpha_{i0}) sin \frac{1}{2}(\alpha_j - \alpha_{j0}) * ...}{cos \frac{1}{2}(\alpha_i + \alpha_{i0} - \alpha_j - \alpha_{j0})} \quad (19)$$

To determine the bounds on each link's orientation error $|\alpha_i - \alpha_{i0}|$, insert Equation (1) into Equation (9) and apply the triangle inequality to obtain:

$$|\alpha_i - \alpha_{i0}| = \left| \sum_{k=1}^{i} \theta_k - \theta_{k0} \right| \leq \sum_{k=1}^{i} |\theta_k - \theta_{k0}| \leq i\delta \quad (20)$$

Applying the inequality of Equation (20) and the simple inequalities

$$cos \theta \leq 1$$
$$\theta \leq |\theta| \quad \forall \theta \in R \quad (21)$$

while keeping in mind that $sin\ \theta$ is a monotonically increasing function for $-\pi/2 < \theta < +\pi/2$, Equation (19) yields

$$\|x_e - x_{e0}\|^2 \leq 4L^2 \left( \sum_{i=1}^{N} sin^2 \frac{i\delta}{2} + \sum_{i=1}^{N} \sum_{j=1, j \neq i}^{N} sin \frac{i\delta}{2} sin \frac{j\delta}{2} \right) \quad \forall i\delta \leq \frac{\pi}{2} \quad (22)$$

Rearranging the right hand side of Equation (22) into polynomial form yields:

$$\|x_e - x_{e0}\|^2 = 4L^2 \left( \sum_{i=1}^{N} sin \frac{i\delta}{2} \right)^2 \quad \forall i\delta \leq \frac{\pi}{2} \quad (23)$$

Thus, for sufficiently small δ, taking the square of Equation (23) yields the root mean square of the end effector as a function of δ:

$$\|x_e - x_{e0}\|_2 \leq 2L \sum_{i=1}^{N} sin \frac{i\delta}{2} \quad (24)$$

**Corollary 1:** For planar robots where the *N*th joint is used to set the orientation *Θ* of the endpoint and the first *N-1* joints are used to set its position, it follows directly from the above analysis that the respective error bounds on position and orientation are given by:

$$\|x_e - x_{e0}\|_2 \leq 2L \sum_{i=1}^{N-1} sin \frac{i\delta}{2} \quad (25)$$
$$|\Theta_e - \Theta_{e0}| \leq N\delta$$

## V. EXAMPLES OF ROBOTS WITH THREE DEGREES OF RECONFIGURABILITY

We demonstrate the construction of an approximation curve for two relatively simple MASR with three revolute joints: one with two actuators and one with one actuator. Each link is 10cm long and 1 cm thick. The MASR is tasked with translating from point A to point B, moving a cup upright along a line, drawing the letter Z, and drawing a circle. The trajectory of the robot must be satisfied within the given error radius of the coordinate space of a fully actuated robot.

### A. Robot moving tip from point A to point B

The simplest task possible for a robot is to move its end-effector from one point to another. The c-space trajectory of the 3DOF robots performing that task is shown in Figure 3. The trajectory in c-space can take any form that starts at the initial coordinate $\theta_A$ and ends at the final coordinate $\theta_B$. It must be emphasized that angles in c-space can affect both position and orientation of the endpoint. Assuming that all maximum angular velocities are $\omega$ and each angle rotates independently, the time of traversal is simply

$$T = \frac{\max_{i \in [1, N]}(|\Delta \theta_i|)}{\omega} \quad (26)$$

For a 2-actuator robot, the trajectory is confined to a series of two-dimensional planes described by Equation (5). These planes constitute the surface of the blue box shown in Figure 3. The axis that span the plane represent the joints that are actuated during the traversal. For example, the right side of the cuboid in Figure 3 is spanned by $\theta_2$ and $\theta_3$; any c-space trajectory on this plane means that the robot is actuated at joints 2 and 3. Traversing across three dimensional c-space entails the trajectory traversing at least two planes, i.e. it must cross at least one boundary between to planes. Denoting *k* as the joint angle that retains an actuator during both phases of actuation, the time is thus given by

$$T = \frac{\max(|\Delta \theta_k|, |\Delta \theta_i|) + \max(|\Delta \theta_k|, |\Delta \theta_j|)}{\omega} + \frac{L}{V} + 2T_{DELAY}$$
$$i, j, k \in \{1, 2, 3\} \quad (27)$$

since there is only one pair of actuator translations.

The one-actuator robot is confined to a series of one-dimensional planes, i.e. lines. These lines constitute the edges of the blue box in Figure 3. The time of traversal would be given by Equation (6).

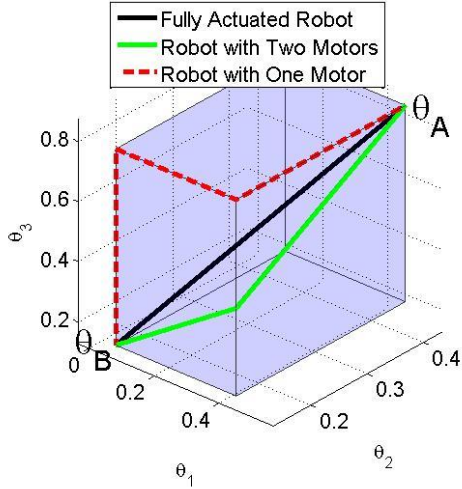

Figure 3. The trajectory of the robots to traverse from initial configuration $\theta_A$ to final configuration $\theta_B$ in three dimensional c-space. The trajectory of the fully actuated robot is represented by the black line. It has no spatial constraints. The trajectory of the 2-actuator robot, shown by the green line segments, is confined to the surfaces of the hypercuboid shown in blue, and that of the 1-actuator robot, shown by the red line segments, is confined to its edges.

## B. Robot orientation and position

This planar robot has three degrees of freedom: two for location and one for orientation. Its task is to move a glass of water along a straight line while keeping it upright. The actuator translates along the robot links, alternating between the position-setting joints (joints 1&2) and the orientation-setting joint (joint 3). A time-lapse snapshot of the robot is shown in Figure 4. The trajectory of a fully-actuated serial robot in *c*-space that moves the cup is represented by the blue dotted curve in Figure 5. We set $\delta = 0.1$ rad.

Following the aforementioned procedure, the trajectory for the MASR with two actuators is shown by the sequence of diagonal green line segments on the surface of the cuboids, while that of the MASR with one actuator is shown by the sequence of straight red line segments on the edges of the cuboids. Each segment is confined to the two dimensional surface of its respective cuboid. These cuboids are constructed in Step 3 of APPROXIMATION-CURVE. Every time that the trajectory moves onto a different face or cuboid, one of the actuators commutes to a different joint. The MASR effectively tracks the fully actuated robot, ensuring that the maximum deviation of corresponding joint angles between the two is never more than $\delta$.

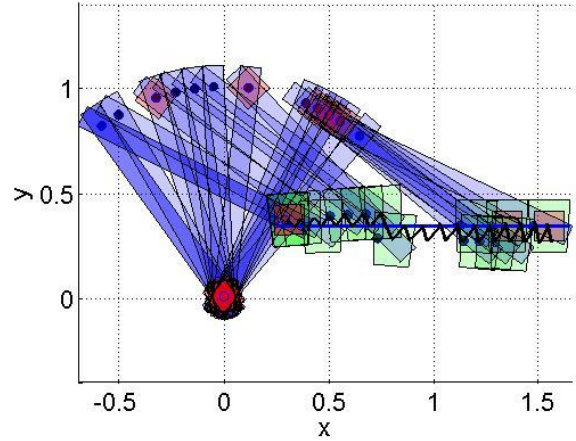

Figure 4. Snapshot of MASR robot transporting a cup along the blue line shown. The actuator, represented by the red rectangle, translates from joint to joint. The trace of the end-effector's trajectory is shown by the black line. For all figures in this article, the units are normalized by the length of a single link.

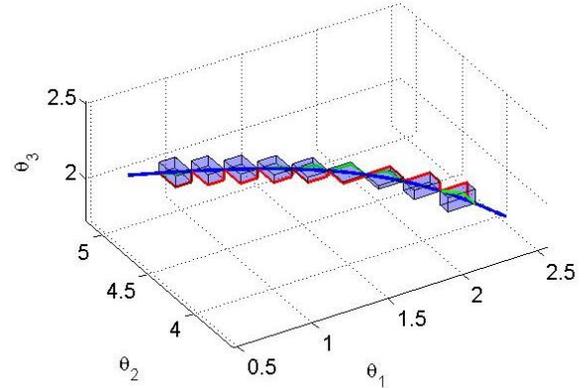

Figure 5. Trajectory of the MASR robot moving a cup in configuration space. The blue dotted line is the trajectory of the fully actuated robot. The limited actuation of the MASR robots results in the constraint on the MASR trajectory in c-space; for any given set of actuator locations, the trajectory is confined to the surface (two actuators-green line segments) or edges (one actuator- red line segments) of the cuboids shown in blue.

The end-point error norm defined by Equation (12) is only relevant when both the MASR endpoint $x_e$ and the fully actuated endpoint $x_{e0}$ are parametrized by the same independent variable yielding a one-to-one correspondence. However, such a parametrization is not necessary; the endpoint error norm may be defined in a similar manner to the c-space error of Definition 1. Using the notation of Definition 1, we denote the parametrized respective endpoints as $x_e(t)$ and $x_{e0}(u)$. The endpoint error norm $\Delta$ is defined as the largest of the distances between the closest distances between any two points on $x_e(t)$ and $x_{e0}(u)$:

$$\Delta : C^1 \times C^1 \to R^+ = \max_t \left( \min_u \left( \| x_e(t) - x_{e0}(u) \|_2 \right) \right) \quad (28)$$

Similarly, the orientation error at the end effector, being by definition the sum of the angular differences, is given by

$$\left|\Theta_e - \Theta_{e0}\right| = \left|\sum_{i=1}^{N} \theta_i - \theta_{i0}\right| \tag{29}$$

This error norm, along with the limit on the error norm of Equation (25), are shown in Figure 6. This validates the analysis of Section IV – the figure clearly demonstrates that the actual error is always less than the error bound, although the gap between them grows with increasing $\delta$.

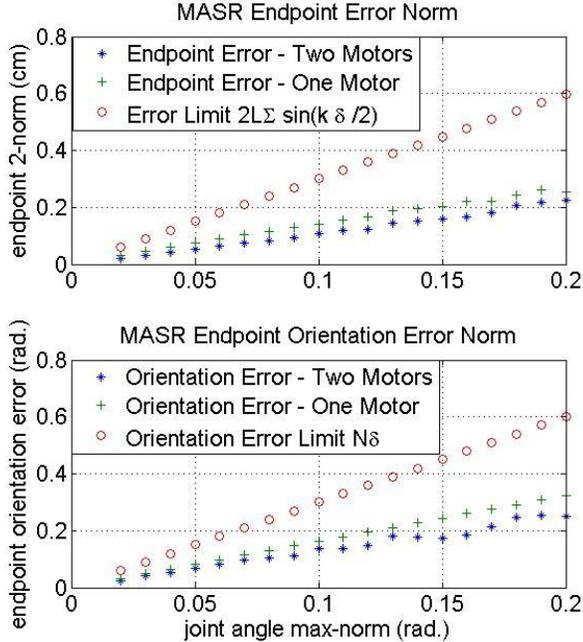

Figure 6. The robot end-effector error $\Delta$ (plus sign & asterisks) and the calculated error limit (circles) as a function of the joint angle limit for both position error (top) and orientation error (bottom). As expected, the actual error is below its maximum possible.

Because the latter has three revolutionary joints while only two endpoint coordinates $x,y$, it has one redundant DOF. There are many different techniques for resolving joint redundancy and different objectives for their resolution. However, the method we select to resolve this redundancy is by selecting the joint angles so as to maximize the determinant of $J^T J$ while constraining the endpoints to stay on target, where $J$ is the Jacobian. This method is chosen because it is a standard objective in robotics that yields the maximum manipulability, or the ability to exert any desired motion at the manipulator's end effector. This was accomplished using the fmincon© function in the MATLAB™ Optimization Toolbox.

### C. Robot drawing the letter Z

The output of the robots' end effectors in tracing the letter Z is shown in Figure 7. A snapshot of the MASR drawing is shown in Figure 8. For robot applications where the end effectors are tasked with tracing a path, this result has significant implications for the selection of actuators of the MASR. The endpoint error of the MASR robot compared with its theoretical limit given by Equation (24) is presented in Figure 9.

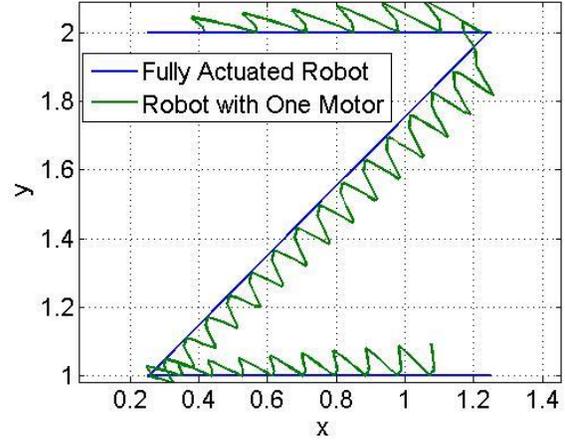

Figure 7. Output of the end-effectors of the MASR robot attempting to draw the letter Z under maximum angular deviation of $\delta = 0.2$ rad. As the figure demonstrates, the end-effector deviation for the single actuator robot is greater than for the two-actuator robot, even though both are bounded by $\delta$.

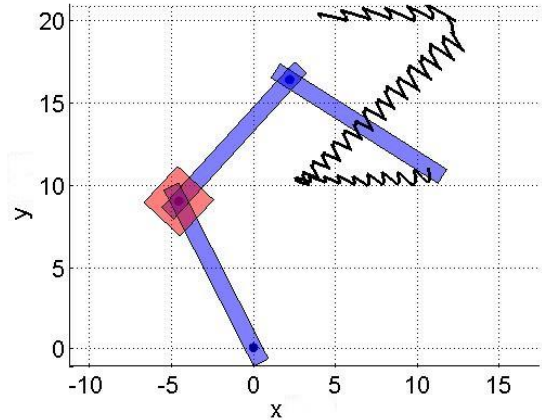

Figure 8. Snapshot of MASR robot drawing the letter Z. The actuator, represented by the red rectangle, translates from joint to joint.

A planar robotic task can be achieved with a minimum of two links and two revolutionary joints. It thus may appear at first glance that having two movable actuators running along three links is an unnecessary complication. However, the extra degree of redundancy is necessary for enabling the robot to navigate around obstacles. In addition, the three DOF provide the robot with extra maneuverability and dexterity that cannot be achieved with a 2DOF robot. Most importantly, the three link robot is mainly a proof of concept for larger hyper-redundant robots with many degrees of freedom.

The effect of lowering the c-space error radius on the number of actuator traversals in drawing the letter $Z$ is shown in Figure 10. As expected, the tighter the error bound is, the more the actuators must switch between the joints of the MASR. As there are two surfaces and three edges between opposite edges of a three dimensional cuboid, the number of traversals for the single actuator MASR will always be 50 percent more than that of the double actuator MASR. This is because each cuboid entails two traversals for the latter, while three for the former.

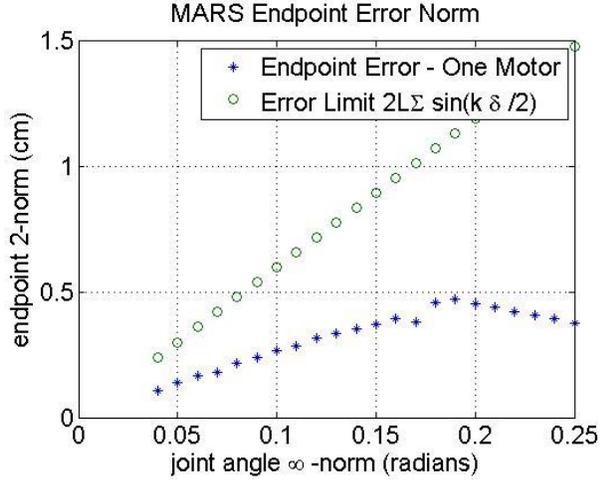

Figure 9. The robot end-effector error Δ (asterisks) and the calculated error limit (circles) as a function of the joint angle limit. As expected, the actual error is below its maximum possible. The actual error for the single motor robot decreases past a certain error norm because the deviation of the robot from its trajectory places it closer to other points along the fully actuated robot's trajectory. Thus, using the definition of Equation (28) to describe the endpoint error may not be the most useful definition.

If the total time for traversal is measured, rather than just the number of actuator shifts, then a similar picture emerges. Assuming a very simple kinematic model where the rotation consumes a constant time per radian $t_r$ and a constant time per actuator translation $t_s$, the total time consumed is given by Equation (6). The total time for the robot drawing the figure Z is also shown in Figure 10, where $t_r$ is taken to be 1.0 seconds per radian and $t_s$ is 1.0 seconds. Here too, the time consumption sharply increases for increasingly small error radii.

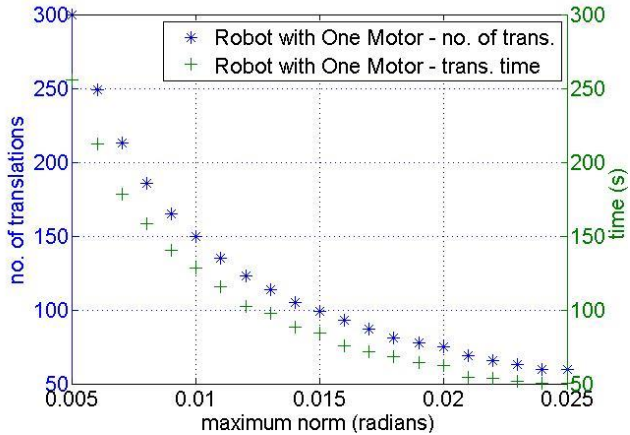

Figure 10. Number of actuator traversals and total time required for the MASR robot to transport an upright cup while remaining within the c-space norm.

### D. Robot drawing a circle.

The results of the same MASR robot drawing a circle is shown in Figure 11. The circle has a radius of 2cm and its origin is located at (10cm, 20cm) from the robot base. Because the task workspace for the circle is smaller than that of the Z, the respective MASR error norm must be correspondingly smaller. The outline drawn in Figure 12 is the result of $\delta = 0.01$ radians. The number of actuator traversals and total time required to draw the circle are shown in Figure 11. Once again, the smaller the error bound is, the more translations are required.

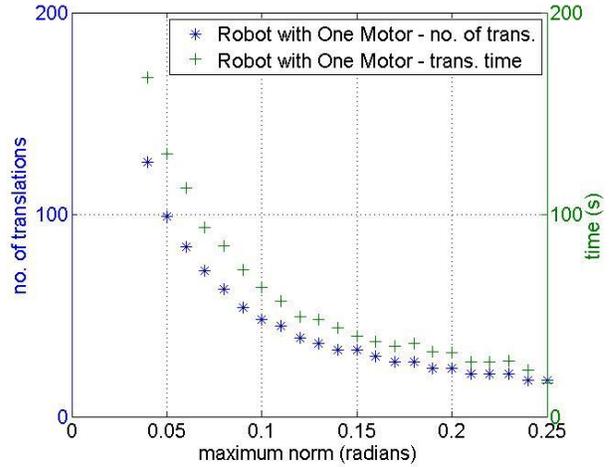

Figure 11. Output of the end-effectors of the MASR robot attempting to draw a circle under maximum angular deviation of $\delta = 0.01$ rad.

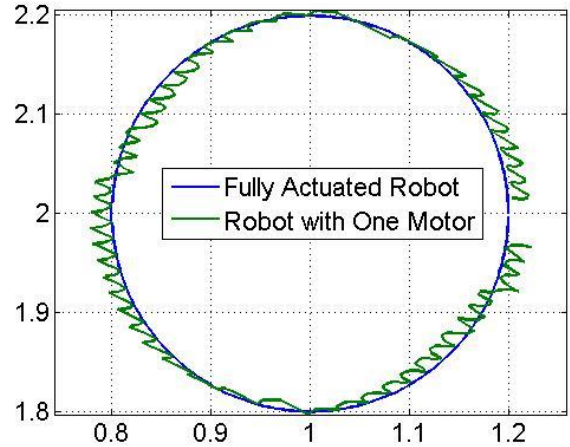

Figure 12. Number of actuator traversals and total time required for the MASR robot to trace a circle while remaining within the c-space norm.

## VI. EXAMPLES WITH HIGHLY REDUNDANT CONFIGURATIONS

To demonstrate the capabilities of the MASR, we simulate a motion planning situation with obstacles as summarized in Figure 13. In this section, the planning was performed by the human operator. The MASR in this example consists of a base and ten links and joints (10 DOF) actuated by one mobile actuator. The goal of the robot is to grab the blue circle and bring it back to the robot's original configuration.

The task is composed of two main challenges. The first is going through the narrow pass of 15 mm, and the second is reaching the target with the small section of the robot that went through the opening. Throughout the whole task, the robot must avoid colliding with the obstacles.

The robot accomplishes this task by having the motor translate and adjust the angles of the joints one at a time. The robot first passes through the narrow pass by transforming its second half into an arc like shape. Then, the mobile actuator passes through the pass and then rotates the links to reach the target. Since four joints and links went through the pass, the robot had four degrees of freedom to reach its target (only three are required in a 2D space to reach location and orientation). In total, only eight translational steps for the motor are required in each direction, demonstrating the dexterity and maneuverability of the MASR.

TABLE I.  MOTION SUMMARY OF MASR. DURING EACH ACTION, THE MOBILE ACTUATOR ROTATES A SPECIFIC JOINT BY AN ANGLE $\theta$ OR ADVANCES FROM JOINT (START) TO ANOTHER (END).

| STEP | Turning [degrees] (joint/angle) | Translation (start-end) |
|---|---|---|
| **Reaching the target** | | |
| 1 | +45 | (1-1) |
| 2 | +45 | (1-2) |
| 3 | -45 | (2-6) |
| 4 | -45 | (6-7) |
| 5 | -45 | (7-9) |
| 6 | -45 | (9-2) |
| 7 | +45 | (2-9) |
| 8 | +30 | (9-10) |
| **Returning after grasping** | | |
| 9 | -75 | (10-10) |
| 10 | -60 | (10-9) |
| 11 | +90 | (9-2) |
| 12 | +30 | (2-10) |
| 13 | +30 | (10-9) |
| 14 | +45 | (9-7) |
| 15 | +45 | (7-6) |
| 16 | -45 | (6-2) |
| 17 | -45 | (2-1) |
| total (absolute) | 840 [degrees] | 48 L |

As shown in Table I, each stage of motion consists of rotating the given joint by the turning angle, then translating the actuator to the desired joint, and repeating the process. There are a total of eight actions required to reach the object, one action to grasp it, and another eight actions required to return to its initial state with the grasped object in hand.

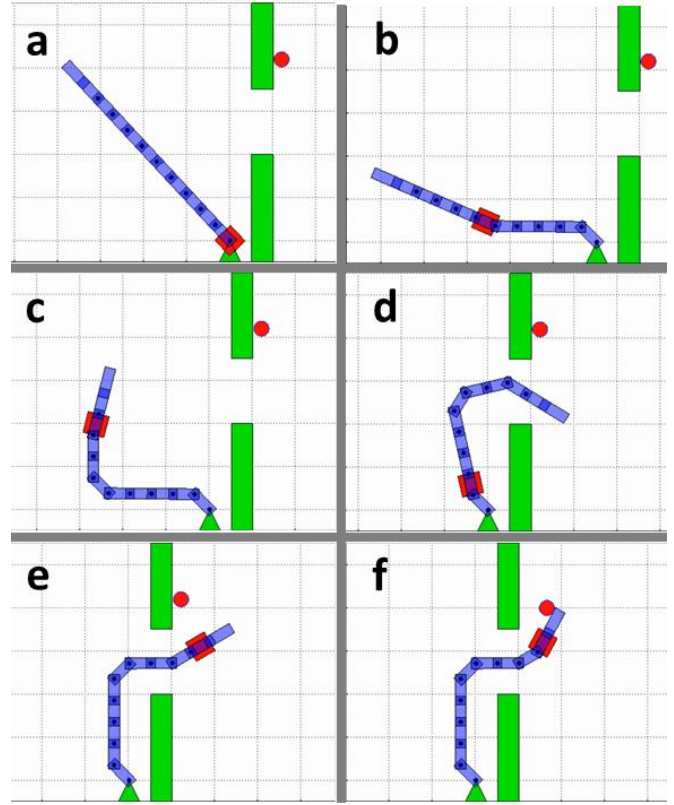

Figure 13. Snapshots of the animation of MASR equipped with a single mobile actuator reaches its target. Starting at (a), the mobile actuator advances to the center after rotating the base link (b). At (c), the mobile actuator rotates the six top links to make an arc shape and then returns to the base (d) to rotate the links and penetrate through the small cavity. The actuator travels again to the top links to rotate them towards the target (e). After reaching its target, the robot makes the inverse plan of a-b-c-d-e to return to its original configuration (f).

The bottom row of Table I shows that the sum total of degrees that the links rotate equals 840°, and the actuator translates a total of 48 link-spans. The total time of the maneuver thus equals the time required to perform both modes of action. With optimal motion planning, however, the latter should be reduced to its minimum possible. Based on Eq.(6), the time required for the locomotion is

$$t_{TOTAL} = \frac{48L}{V} + \frac{840}{\dot{\theta}} + 17\, t_{DELAY} \qquad (30)$$

## VII. EXPERIMENTS

### A. Robot design

To prove the feasibility of MASR, we designed a manufactured a mobile actuator, with 10 links and a base. The robot parts are 3D printed using Object Connex 350 with nominal accuracy of nearly 50 microns using "Verogray" material. In this version, the joint angle is passively locked by a spring applying a friction force. To increase the friction force we glued sand papers to the links and inserted a metal screw to the clamp. At their bottom, the links have a track which allows the mobile actuator to travel along them to reach and actuate a desired joint. Each of the links is 2 cm wide and 5 cm long, giving the active section of the snake robot a total length of 50 cm. The weight of the mobile actuator is 102 grams, whereas the average weight of a links including the clamp and joint is nearly 25 grams. We attached a magnet to the tip of the last link in order to grasp our target. However, other grasping mechanisms can be added.

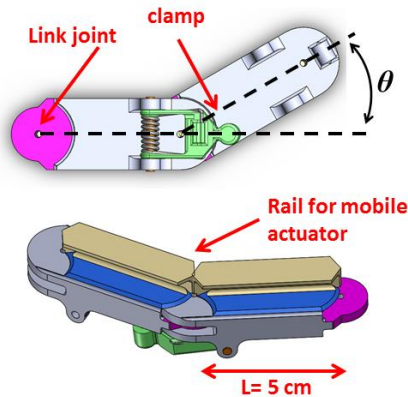

Figure 14. A top and bottom view of two adjacent links. The relative orientaion between the links is passively fixed by the clamps.

The mobile actuator (presented in Figure 15) has two motors. One motor actuates the wheels to drive the mobile actuator along the tracks of the links, and a second motor to rotate the links. The rotational motor is attached to a linear gear mechanism, allowing the teeth to disconnect from the links or push them for rotation. The maximum relative angle between the links is 45 degrees. We used a 4 Volts

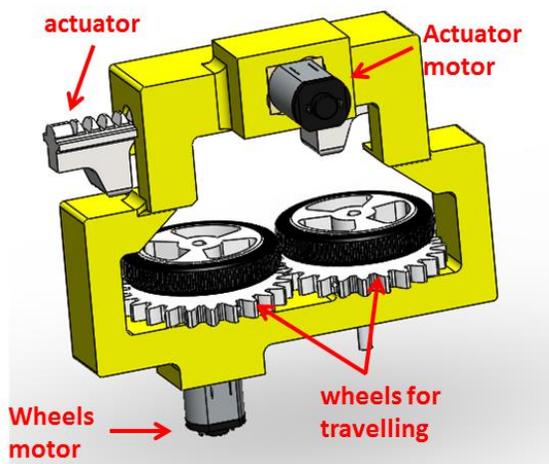

Figure 15. The mobile actuator that travels upon the links. This actuator has two motors, one motor to travel along the links and a second motor to rotate the links.

Lithium-ion battery to actuate the motors. The speed of the locomotion is nearly 3 cm/s and the rotational speed is nearly 18 degrees/s. The robot is very modular and the number of mobile actuators and links is easily changeable. We used motors with 1000:1 gear ratio which can produce 0.9 Nm of torque at 32 rpm. This torque is necessary to overcome the friction torque between the different links and other external forces to produce motion.

During all of the experiments, the mobile actuator was remotely controlled by a human operator. The operator had a two channel joystick. One channel is used to drive the mobile actuator forward and backward along the links and the other to rotate the links clockwise or counter clockwise.

Note that in this preliminary prototype, there is no locking/unlocking mechanism (which we believe will result in superior performance in terms of accuracy and loads). Rather, the system is passively locked with friction and the mobile actuator fitted with a strong motor overcomes the friction to rotate the links.

### B. Experiments with 5 links

In order for the robot to operate as planned, it must be able to perform the following mechanical operations:
1. Travel freely over the links forward and backward.
2. Travel over curved joints without changing their orientation. (the links are passively locked)
3. Rotate the links.

The basic experiment is presented in Figure 16. The mobile actuator was tested going towards the end of the links and returning back with and without bending the links. In both cases, the robot had no difficulty travelling over the links or rotate them to either direction.

Starting at (a), the robot advances towards its tip (b-c), then returns to the center (d). The robot then rotates the links clockwise (e) and counter clockwise (f). The robot then travels over the curved joint (g) and rotates its tip clockwise (h) and counter clockwise (i). The robot then moves to the tip (j) (see movie).

As the joints can be rotated by 45 degrees to each direction, the robot can make a c shape (half a circle) by rotating 4 links in the same direction (counter clockwise). This experiment is illustrated in Figure 17 (see movie).

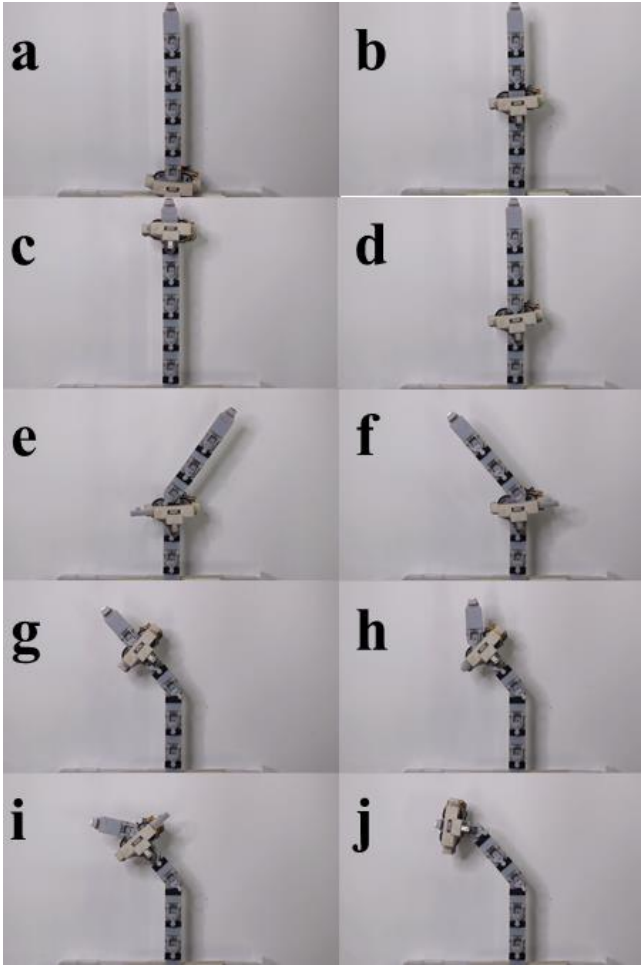

Figure 16. The mobile actuators travel forward and backward over the links without changing their orientation and activate them to the desired location.

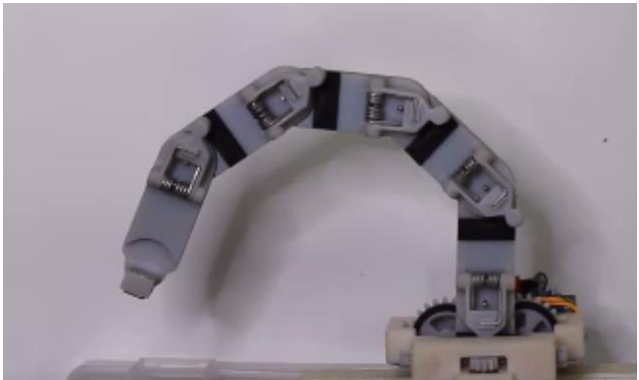

Figure 17. By rotating the four links counter clockwise, the robot gets a C shape.

### C. Experiments with 10 links

In the following experiment, we added 5 more links to the robot (10 in total). The robot is very modular and adding the links requires nearly 2 minutes. With the longer version, we performed a task that is similar to the example presented in Section IV. The results are presented in

Following the same algorithm, the robot successfully reached its desired target. However, we found that since the robot is made of printed material, it slightly cured downwards by nearly 1 cm. Even though the weight of the robot is larger and the torque acting on the links substantially increased, the links remained locked during the experiment.

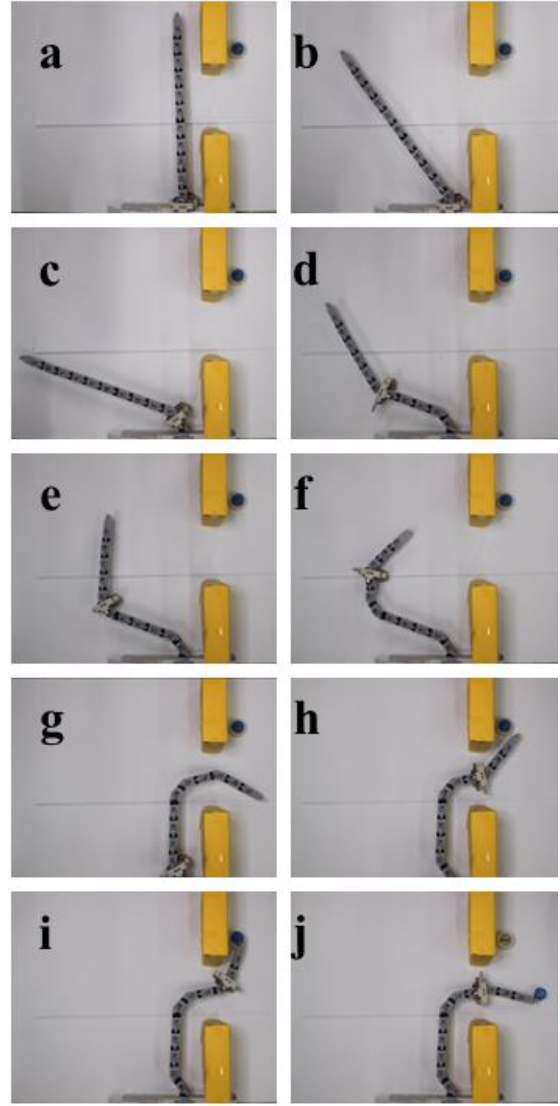

Figure 18. The robot penetrating through a small pass to reach a target being the wall.

## VIII. SUMMARY AND CONCLUSIONS

This paper has introduced a minimally actuated robotic snake (MASR). The MASR can execute complex motions with a small number of actuators. It consists of a mobile actuator that shifts its position along the joints of the robot. This enables it to shape the robot to any desired position by incrementally adjusting all of its joints. This was shown by an example of where it successfully manipulates an object while maneuvering around obstacles. We have described the unique kinematics of the MASR and demonstrated how it can duplicate the motion of a fully actuated robot to within any desired degree of accuracy.

The robot is suitable for applications in a complex and confined environment with low payload and that do not require rapid deployment. While the robot cannot hold large weights, it is a "rigid" mechanism (not compliant) in the sense that it is not meant to deform due to performance of its tasks. The robot is also very modular - the number of links and mobile actuators can be changed in a matter of minutes.

We built an experimental robot with 10 links and one mobile actuator. We used the robot to show how by using a single mobile actuator, it is possible to control the 10 joints of our robot and penetrate through a confined space and reach the target. We found that the control is simple and intuitive, and only a few minutes are required for a human operator to learn how to actuate the robot. We were able to perform the tasks that included going through a small pass and reaching a target. The robot can achieve different configurations as c shape or s shape.

Further research and development of the MASR is ongoing. New improved designs are being developed for the physical actuating mechanism that will yield more rigid structure (by producing metal links), smoother motions, and reduce errors and malfunctions by fitting the mobile actuator with a controller and sensors.

In our future work we aim at developing a comprehensive general motion planning algorithm to yield optimal motions for the MASR in an obstacle environment for one or more actuators.